%% file: PaperForReview.tex
\documentclass[10pt,twocolumn,letterpaper]{article}

\usepackage[pagenumbers]{wacv} 

\usepackage{graphicx}
\usepackage{amsmath}
\usepackage{amssymb}
\usepackage{booktabs}

\usepackage[pagebackref,breaklinks,colorlinks]{hyperref}

\usepackage[capitalize]{cleveref}
\crefname{section}{Sec.}{Secs.}
\Crefname{section}{Section}{Sections}
\Crefname{table}{Table}{Tables}
\crefname{table}{Tab.}{Tabs.}

\usepackage{mathtools}
\usepackage{amsmath}
\usepackage{enumitem}
\usepackage{colortbl}
\usepackage{pifont}
\definecolor{gray}{rgb}{0.85,0.85,0.85}

\newcommand{\name}{\texttt{MOOSS}}
\input{math_commands}

\def\wacvPaperID{465} 
\def\confName{WACV}
\def\confYear{2025}

\begin{document}

\title{\texttt{MOOSS}: Mask-Enhanced Temporal Contrastive Learning for Smooth State Evolution in Visual Reinforcement Learning}

\author{Jiarui Sun, M. Ugur Akcal, Girish Chowdhary\\
University of Illinois Urbana-Champaign\\
Urbana, IL, USA\\
{\tt\small \{jsun57, makcal2, girishc\}@illinois.edu}
\and
Wei Zhang\\
Visa Research\\
Foster City, CA, USA\\
{\tt\small wzhan@visa.com}
}
\maketitle

\input{files/abs}

\input{files/intro}
\input{files/related}
\input{files/prelim}
\input{files/method}
\input{files/exp}
\input{files/conc}

\input{files/ack}

{\small
\bibliographystyle{ieee_fullname}
\bibliography{egbib}
}

\normalsize{
\newpage
\input{files/app}
}

\end{document}

%% file: math_commands.tex
\def\eqref#1{equation~\ref{#1}}

\def\1{\bm{1}}

\def\rva{{\mathbf{a}}}

\def\rvk{{\mathbf{k}}}

\def\rvo{{\mathbf{o}}}
\def\rvp{{\mathbf{p}}}
\def\rvq{{\mathbf{q}}}

\def\rvs{{\mathbf{s}}}

\def\rmW{{\mathbf{W}}}

\def\gA{{\mathcal{A}}}

\def\gE{{\mathcal{E}}}

\def\gG{{\mathcal{G}}}

\def\gK{{\mathcal{K}}}
\def\gL{{\mathcal{L}}}

\def\gO{{\mathcal{O}}}

\def\gS{{\mathcal{S}}}

\def\gV{{\mathcal{V}}}

\def\sR{{\mathbb{R}}}

\newcommand{\E}{\mathbb{E}}

\newcommand{\R}{\mathbb{R}}



%% file: files/abs.tex
\begin{abstract}
  In visual Reinforcement Learning (RL), learning from pixel-based observations poses significant challenges on sample efficiency, primarily due to the complexity of extracting informative state representations from high-dimensional data. 
  Previous methods such as contrastive-based approaches have made strides in improving sample efficiency but fall short in modeling the nuanced evolution of states. 
  To address this, we introduce \name, a novel framework that leverages a temporal contrastive objective with the help of graph-based spatial-temporal masking to explicitly model state evolution in visual RL. 
  Specifically, we propose a self-supervised dual-component strategy that integrates (1) a graph construction of pixel-based observations for spatial-temporal masking, coupled with (2) a multi-level contrastive learning mechanism that enriches state representations by emphasizing temporal continuity and change of states.
  \name~advances the understanding of state dynamics by disrupting and learning from spatial-temporal correlations, which facilitates policy learning. 
  Our comprehensive evaluation on multiple continuous and discrete control benchmarks shows that \name~outperforms previous state-of-the-art visual RL methods in terms of sample efficiency, demonstrating the effectiveness of our method.
\end{abstract}

%% file: files/intro.tex
\section{Introduction}
\label{sec:intro}

Visual Reinforcement Learning (RL), \ie, an RL agent learning from visual signals composed of sequences of image-based observations, has long been a significant challenge. 
Compared to RL that utilizes compact state-based features, Visual RL is notably \textit{sample inefficient}: it requires more environment interactions for a visual RL agent to achieve a comparable performance to its state-based counterparts \cite{tassa2018deepmind}. 
This inefficiency primarily stems from the complexity in extracting informative states from high-dimensional visual data (pixels).
Despite this, visual RL's ability to function without handcrafted features offers broad applicability and a close resemblance to natural learning processes.
Therefore, the ability to efficiently learn effective state representations is crucial.

\begin{figure}[t]
\centering
\includegraphics[width=0.8\linewidth]{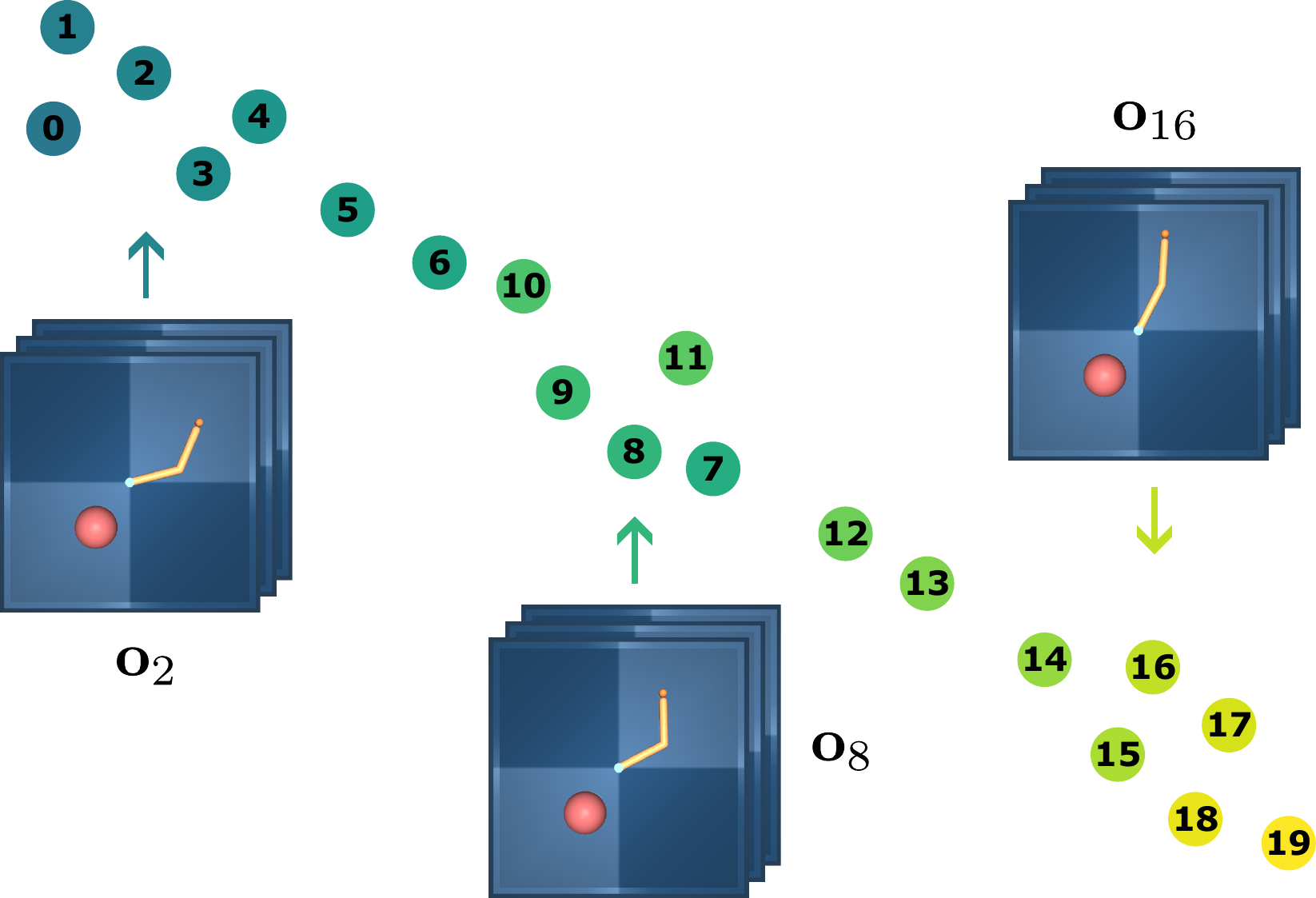}
\caption{t-SNE \cite{van2008visualizing} visualization of the state representations from a trained visual RL agent on the \textit{reacher-easy} task from DeepMind Control Suite \cite{tassa2018deepmind}. 
The state representations are encoded from an observation sequence $\rvo_{0:19}$ of length 20, guided by random actions. 
Numbers within the color-coded dots denote the temporal indices.
Note that the t-SNE visualization demonstrates a temporal order, suggesting a gradual, smooth evolution of the states.
}
\label{fig:tsne}
\end{figure}

To this end, many approaches improve sample efficiency of visual RL agents through incorporating auxiliary tasks tailored to benefit the learning of informative state representations.
These auxiliary tasks often rely on \textit{self-supervision} signals, which are derived from trajectory roll-outs obtained from agent-environment interactions. 
Examples of these tasks include learning forward \cite{schwarzer2020data} or backward \cite{paster2020planning} predictive features, predicting rewards \cite{shelhamer2016loss}, and applying bisimulation metrics \cite{zhang2020learning}.
Among numerous ways to facilitate state representation learning, \textit{contrastive-based} approaches have emerged as a prominent framework, focusing on maximizing agreement between different views of a state.
For example, CURL \cite{laskin2020curl} generates positive samples of state through image augmentation techniques;
subsequent works such as ATC \cite{stooke2021decoupling} treat encoded observations separated by a short temporal difference as positive samples, introducing the temporal concept to the contrastive objective.
On the other hand, methods involving masked reconstruction, such as MLR \cite{yu2022mask}, which perform reconstruction from corrupted observations, are less common yet offer unique insights.
These auxiliary objectives have shown great improvements in sample efficiency for visual RL.

However, the effectiveness of current methods is limited by their inadequate consideration of \textit{state evolution}.
Specifically, if we consider observations or states within adjacent timesteps, as exampled in \cref{fig:tsne}, it becomes apparent that they typically exhibit stronger temporal correlations, \ie, more ``similar'', due to their inherent causal relationships, as opposed to those further apart.
This suggests that state embeddings, encoded from raw observations, are likely to evolve temporally in a gradual and smooth manner, with abrupt changes being less probable.
However, existing contrastive methods only consider a \textit{binary distinction} between positive and negative samples, overlooking the gradual evolutionary nature of states.
In addition, unlike video models \cite{blattmann2023stable} that can process multiple frames simultaneously to capture temporal evolution, RL's formulation constrains the observation encoder to map \textit{one} observation to \textit{one} state independently.
This makes temporal modeling even harder.
On the other hand, approaches within the masked reconstruction domain often adopt a uniform masking approach, overlooking the high spatial-temporal correlation of consecutive pixel-based observations.
We argue that such reconstruction task does not sufficiently challenge the model to understand the underlying dynamics of the observations, making the learned state representations less informative.
These limitations in both contrastive and masked reconstruction methods – the former's binary view of sample relationships and the latter's oversight of spatial-temporal nuances – impede a deeper understanding of state dynamics, which is essential for progress in efficiency of visual RL.

To address the above limitations, we propose to explicitly model the state evolution for efficient state representation learning via self-supervision.
Our approach, \name, \textbf{M}ask-enhanced temp\textbf{O}ral c\textbf{O}ntrastive learning for \textbf{S}mooth \textbf{S}tate evolution, explores the potential of combining contrastive learning with spatial-temporal mask modeling.
Specifically, as shown in \cref{fig:mdl}, \name~integrates an auxiliary temporal contrastive objective into visual RL agents, which is jointly trained with the main RL objective.
This contrastive objective goes beyond the conventional binary distinction by modeling state similarities at \textit{multiple levels}.
This allows us to encourage the model to focus on gradual and evolving state changes over various temporal distances.
Alongside this, we envision pixel-based observations as a \textit{spatial-temporal graph}, applying a random walk-based masking technique. 
This presents a complex pretext task, posing greater challenges than those presented by standard uniform block-based masking \cite{yu2022mask}, thereby compelling the RL agent to acquire a deeper understanding of observations with deliberately disrupted spatial-temporal connections. 
By combining these approaches, \name~applies the temporal contrastive objective to embeddings from both masked and unmasked observations.
This unified strategy enhances the model's ability to efficiently capture the dynamics of the observations by encouraging the agent to focus on evolving elements, thus facilitating informative state learning and improve policy learning.

Our main contributions are summarized as follows.
(1) We propose a novel, auxiliary temporal contrastive objective tailored to visual RL, aimed at emphasizing the temporal continuity and change of states derived from pixel-based observations.
(2) We re-cast pixel-based observations as a spatial-temporal graph, employing random walk-based masking to generate contrastive samples with disrupted spatial-temporal correlations.
(3) Combining temporal contrastive objective with spatial-temporal masking, we introduce \name. 
\name~is proven effective for improving the sample efficiency of visual RL algorithms across multiple continuous and discrete control benchmarks, including the DeepMind Control Suite \cite{tassa2018deepmind} and Atari games \cite{bellemare2013arcade}, outperforming previous state of the art. 
Our detailed ablation studies further validate the efficacy of our method.

%% file: files/related.tex
\section{Related Work}
\label{sec::related}

\subsection{Representation Learning for Visual RL}

Efficiently learning informative state representations from pixel-based observations is a challenging problem for RL.
Unlike the abundance of data in supervised settings, RL relies on experience trajectories collected through costly agent-environment interactions.
This makes robust observation encoding from limited samples a complex task.
As such, sample efficiency has emerged as a critical focus area for visual RL, with various approaches being developed to address this problem.
Some methods involve learning world models \cite{hafner2019dream, kaiser2019model, hafner2019learning, seo2023masked, pan2022iso}, where the aim is to construct an internal representation of the environment that aids policy learning. 
Few other works \cite{laskin2020reinforcement, hansen2021stabilizing, kostrikov2020image, ma2023learning, huang2022spectrum} emphasize enhancing observation diversity through data augmentation techniques.
Through enriching training samples, these methods acquire observation encoders that are more robust and generalizable, thereby alleviating the efficiency issue. 
Facilitated by data augmentation, one major line of work involves leveraging self-supervised auxiliary objectives that are optimized jointly with policy learning objectives. 
Notable examples include learning forward or backward predictive features \cite{shelhamer2016loss, gelada2019deepmdp, guo2020bootstrap, schwarzer2020data, lee2020stochastic, yu2021playvirtual}, and state reconstruction \cite{yarats2021improving, zhu2022masked, yu2022mask}.
Within state reconstruction methods, MLR \cite{yu2022mask} stands out by performing latent reconstruction from corrupted pixels, marking an early exploration of mask-based modeling in visual RL.

\begin{figure}[!t]
\centering
\includegraphics[width=1.0\linewidth]{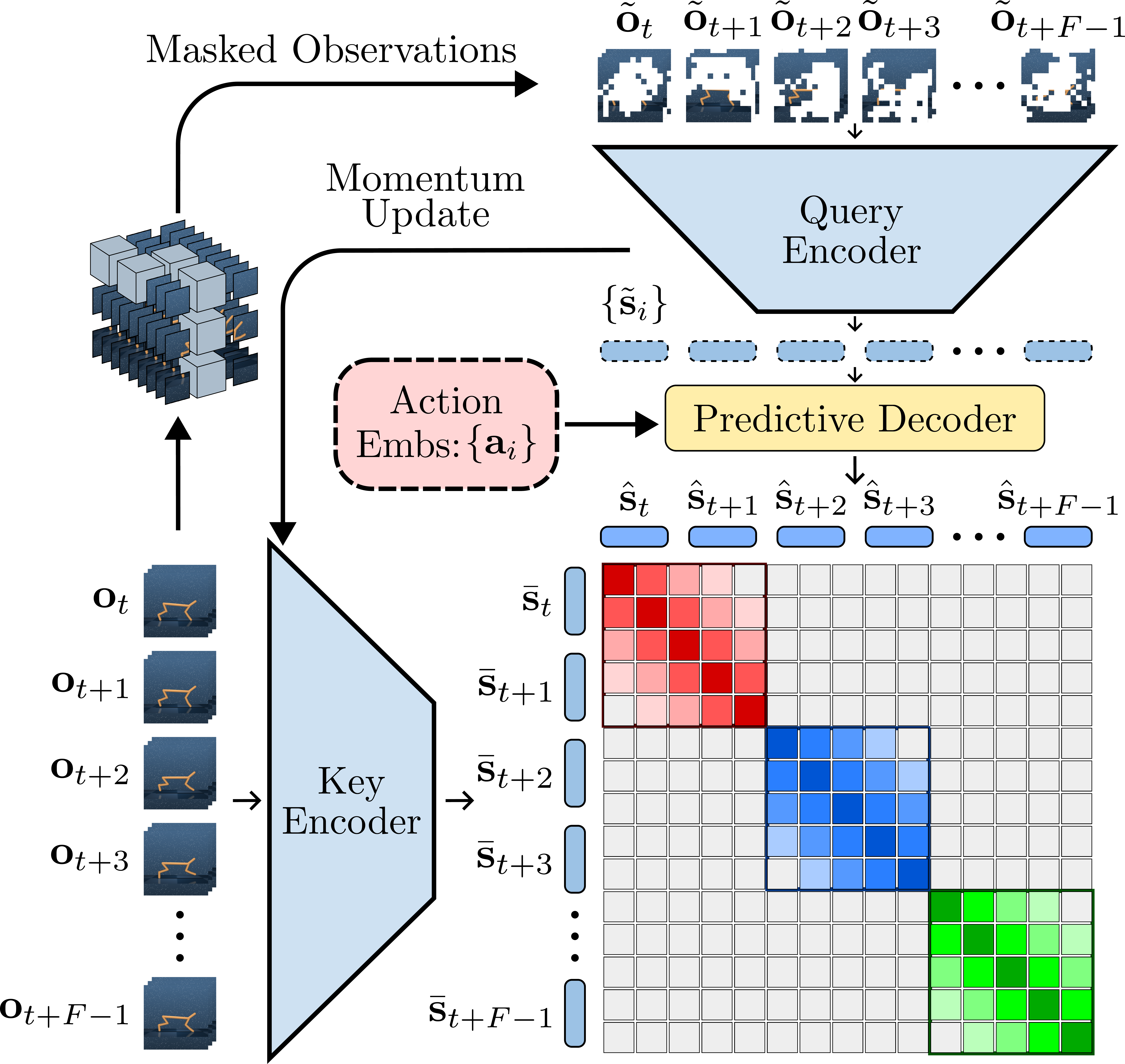}
\caption{The proposed \name~framework. 
We first perform graph-based spatial-temporal masking on the observation sequence $\rvo_{{t}:{t+F-1}}$.
The masked observations are then fed into a query encoder, generating ${\tilde{\rvs}}_i$s.
The unmasked observations are processed by a momentum key encoder.
The key encoder generates the \textit{key state embeddings} $\bar{\rvs}_{{t}:{t+F-1}}$.
A predictive decoder is used to further process the outputs $\tilde{\rvs}_i$s of the query encoder, generating the \textit{query state embeddings} $\hat{\rvs}_{{t}:{t+F-1}}$ conditioned on the corresponding action embeddings ${\rva}_i$s (Embs).
}
\label{fig:mdl}
\end{figure}

Among these auxiliary tasks, contrastive discrimination \cite{laskin2020curl, liu2021return, oord2018representation, mazoure2020deep, stooke2021decoupling, zheng2023taco, anand2019unsupervised} has emerged as a prominent technique for enhancing state representation learning.
The seminal work CURL \cite{laskin2020curl} focuses on maximizing agreement between augmented versions of the same observation.
Subsequent works integrate temporal elements into their contrastive objectives. 
ATC \cite{stooke2021decoupling} and ST-DIM \cite{anand2019unsupervised} treat temporally close neighbors as positive samples to emphasize temporal proximity, whereas DRIML \cite{mazoure2020deep} and TACO \cite{zheng2023taco} focus on aligning predicted future states with their groundtruth counterparts.
In addition to this joint learning scheme, another major direction of research aims to acquire robust, informative state representations from pretrained encoders before policy learning \cite{wang2022vrl3, liu2021aps, liu2021behavior, schwarzer2021pretraining} as a separate stage.
Our approach, \name, falls in the auxiliary joint learning framework, explores the potential of combining contrastive learning with mask modeling to explicitly model state evolution.

\subsection{Contrastive Learning and Masked Modeling}

Contrastive learning, a self-supervised representation learning approach, has gained significant attention and been applied in various fields such as computer vision \cite{chen2020simple, he2020momentum} and graph learning \cite{you2020graph, xu2021infogcl}. 
The most prominent objective in contrastive learning is the InfoNCE loss \cite{oord2018representation}, designed to maximize the mutual information between positive samples. 
Formally, given a query $q$ and a key set $\gK$ containing its positive key $k^{+}$, the objective $\mathcal{L}_{q}$ is to ensure that $q$ aligns more closely with $k^+$ than with other keys in $\gK$:
\begin{equation}
\label{eq::infonce}
    \mathcal{L}_{q}=-\E\left[\log \frac{\exp(\textrm{sim}(q, k^+)/\tau)}{\sum_{{k}\in \gK}{\exp(\textrm{sim}(q, k)/\tau)}}\right],
\end{equation}
where $\textrm{sim}(\cdot)$ measures the similarity of the sample pair, and $\tau$ is the temperature parameter. 
In visual RL, this similarity is typically calculated through a bilinear product \cite{laskin2020curl, stooke2021decoupling, zheng2023taco}.

However, despite various principles are used to form the positive pair $(q, k^+)$, the contrastive objective focuses only one unique positive pair for each query state. 
This approach, while effective, adheres to a binary distinction, categorizing interactions solely as positives or negatives. 
Some works form other fields aim to broaden this perspective by allowing multiple positive samples for one query. 
Approaches such as MIL-NCE \cite{miech2020end} and CoCLR \cite{han2020self} incorporate multiple positive keys to one query into their contrastive loss to learn video representations.
RINCE \cite{hoffmann2022ranking} further extends the binary distinction by preserving a ranked ordering of positive samples, showing effectiveness in supervised classification task with additional superclass labels and unsupervised video representation learning.
Inspired by RINCE, \name~is the first visual RL approach using a multi-level temporal contrastive objective to model state evolution.

Masked modeling, with roots dating back to \cite{vincent2008extracting}, has recently gained prominence in language \cite{devlin2018bert, sun2019ernie}, vision \cite{bao2021beit, he2022masked}, and graph \cite{tan2023s2gae, hou2022graphmae} domains. 
Its effectiveness in training models through self-supervised reconstruction has made it a preferred choice for many studies.
While reconstruction has proven to be a powerful pretext task, masking techniques vary significantly among domains. 
Language models typically perform masking at the token level, obscuring specific words or phrases to encourage the model to predict the missing information based on context.
Image models often employ patch masking \cite{he2022masked, gao2022convmae} due to the heavy spatial redundancy of images, while some video models utilize techniques such as tube masking \cite{tong2022videomae, wang2022bevt} to incorporate the temporal dimension.
For graph learners, strategies range from uniform \cite{hou2022graphmae} to path-based \cite{li2023s, sun2023revealing} masking.
In our work, we explore the application of graph masking principles to image-based observation sequences in visual RL. 
Through experiments, we demonstrate that this creates a challenging pretext task, compelling \name~to develop a deep understanding of state dynamics and enhancing its ability to interpret complex spatial-temporal patterns of visual data.

%% file: files/prelim.tex
\section{Preliminaries}
\label{sec::prelim}

The learning process of Visual RL corresponds to a Partially Observable Markov Decision Process (POMDP) \cite{bellman1957markovian, kaelbling1998planning}: $(\gO, \gA, P, R, \gamma)$, where $\gO$, $\gA$, $P$, $R$, $\gamma$ denote the observation space, the
action space, the transition dynamics $\gO \times \gA \rightarrow \Delta(\gO)$, the reward function $\gO \times \gA \rightarrow \sR$, and the discount factor, respectively. 
$\Delta(\mathcal{O})$ is the space of probability distributions over $\gO$, and the reward function at time step $t$ can be written as $r_t = R(\rvo_t, a_t)$, where $a_t$ is the $t^{th}$ action.
For visual RL, each observation $\rvo_t \in \R^{c\times H\times W}$ consists of $c$ two-dimensional pixel-based feature maps.
The objective of the RL agent is to learn a policy $\pi(a_t | \rvo_t)$ which maximizes the discounted cumulative reward $\E_\pi \sum_{t=0}^{\infty} \gamma^t r_t$, where $\gamma \in [0,1)$.

%% file: files/method.tex
\section{Methodology}

\begin{figure}[t]
    \centering
    \includegraphics[width=0.6\linewidth]{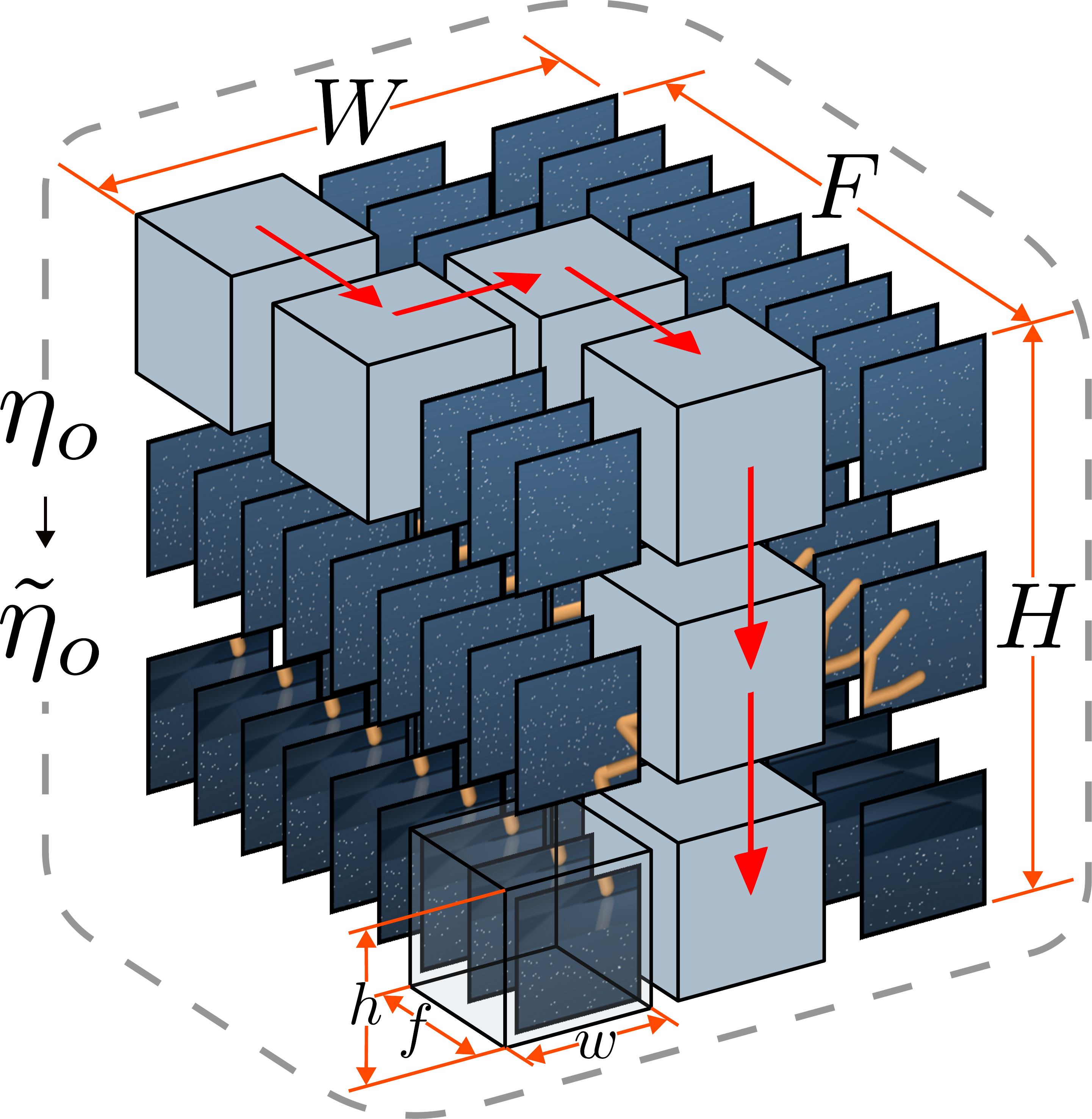}
    \caption{Illustration of our graph-based spatial-temporal masking.
    The observation sequence $\eta_o$ with shape $F \times H \times W$ is equally divided into non-overlapping cubes with shape $f \times h \times w$, constructing a spatial-temporal graph $\gG$ with adjacent nodes connected. Masking is applied by simulating a random walk on the constructed graph.}
    \label{fig:st_mask}
\end{figure}

As a method designed for efficient state representation learning in visual RL, \name~can be seamlessly integrated with any existing RL algorithms, such as SAC \cite{haarnoja2018soft} or Rainbow \cite{hessel2018rainbow}. 
This integration is achieved by combining policy updates from the chosen RL algorithm with \name's auxiliary contrastive loss updates.
The core idea of \name~is to explicitly model state evolution through (1) graph-based spatial-temporal masking on pixel-based observations for contrastive sample generation, and (2) a carefully designed multi-level temporal contrastive objective with the help of the masking approach.
In the following subsections, we first present \name's overall framework, then introduce the proposed masking module with related architectural designs in detail.
We then delve into the specifics of the temporal contrastive objective.

\subsection{Overall Framework}

The \name~framework, illustrated in \cref{fig:mdl}, begins by constructing a spatial-temporal graph from the raw, pixel-based observations. 
On this graph, a masking operation is performed. 
The graph's masked observations, alongside their unmasked counterparts, are then fed into an observation query encoder and a momentum key encoder, respectively, to produce state embeddings. 
The masked state embeddings are then passed to a predictive decoder to generate \textit{query} states, while the unmasked observations are used to form \textit{key} states. 
Finally, the temporal contrastive objective is applied to these query and key state representations, with the aim of modeling the evolution of states over time.

\subsection{Graph-based Masking for State Generation}

\paragraph{Spatial-Temporal Masking.}
We perform graph-based spatial-temporal masking to obtain masked observation sequences which are used to generate the query embeddings.
The masking process is illustrated in \cref{fig:st_mask}.
Let $\eta_o \coloneqq \{\rvo_i\}_{i=t}^{t+F-1}$ denote a sequence of observations with $F$ timesteps sampled from the replay buffer.
We first stack all observations in $\eta_o$ as a cuboid of shape $F \times H \times W$.\footnote{Here we omit the feature dimension $c$ for notation simplicity.}
Then, we equally divide the cuboid into non-overlapping cubes with the shape of $f \times h \times w$, where each cube can be thought of as a node on a graph.
For two such nodes that are adjacent to each other, \ie, two cubes that are spatial-temporally consecutive, we form an edge in between.
As such, we construct a spatial-temporal graph $\gG = (\gV, \gE)$ from the observation sequence.
$\gG$ contains $\frac{FHW}{fhw}$ nodes by construction.

We then randomly mask a portion of the nodes from $\gG$ to obtain a masked observation sequence $\tilde{\eta}_o \coloneqq \{\tilde{\rvo}_i\}_{i=t}^{t+F-1}$.
Instead of uniformly masking image patches as in previous works \cite{yu2022mask}, we propose to use random walk-based masking on the constructed graph $\gG$.
Formally, the set of masked nodes $\gV_{\textrm{mask}}$ with size $|\gV| \cdot p_m$ are collected from a sampled random walk $\gE_{\textrm{mask}}$ as:
\begin{equation}
    \gE_{\textrm{mask}} \sim \textrm{RandomWalk}\left(\gE, r\right),    
\end{equation}
where $p_m$ is the masking ratio, and $r \in \gV$ is the root node to start the walk.
Then, all cubes corresponding to nodes in $\gV_{\textrm{mask}}$ are masked by setting the corresponding patches to zero to form $\tilde{\eta}_o$.
Compared to uniform patch-based masking, our graph-based spatial-temporal masking can more effectively break short-range consecutive information chunks.
As the information density of image-based observation sequences is relatively low due to the spatial-temporal redundancy of visual data, our method creates a more challenging pretext task for the subsequent modules to solve.

\paragraph{Observation Encoding.}
Inspired by works in self-supervised image
representation learning \cite{he2020momentum, guo2020bootstrap}, two observation encoders are used to generate state embeddings from (1) the masked and (2) the original observations, respectively. 
The encoders are Convolutional
Neural Network (CNN)-based, and their architectural design are taken from previous works \cite{yarats2021improving, tassa2018deepmind}.
First, one encoder $f_\theta(\cdot)$ is used to process $\tilde{\eta}_o$, which generates a sequence of masked state embeddings $\tilde{\eta}_s \coloneqq \{\tilde{\rvs}_i\}_{i=t}^{t+F-1}, \Tilde{\rvs}_i \in \sR^d$. 
The parameters of $f_{\theta}(\cdot)$ are optimized in an end-to-end manner.
At the same time, another momentum observation encoder $f_{\bar{\theta}}(\cdot)$ is used to encode the original observations $\eta_o$ to produce the \textit{key state embeddings} $\eta_k$:
\begin{equation}
    \eta_k \coloneqq \{\bar{\rvs}_i\}_{i=t}^{t+F-1} = f_{\bar{\theta}}(\eta_o).
\end{equation}
This second encoder $f_{\bar{\theta}}(\cdot)$ shares the same architecture as $f_{\theta}(\cdot)$, and its parameters $\bar{\theta}$ are updated by an Exponential Moving Average (EMA) of $\theta$ with the momentum coefficient $m \in [0, 1)$ as $\bar{\theta} \leftarrow m \bar{\theta} + (1-m) \theta$.

\paragraph{Predictive Decoding.}
RL naturally operates sequentially: an agent's current state is determined by its past states and actions.
Thus, the actions stored in the trajectory roll-outs provide crucial guidance in state evolution.
Considering this, we utilize both states and actions as the inputs to a causal Transformer-based predictive decoder for query state generation, reducing possible ambiguities to facilitate the subsequently described temporal contrastive objective.
Formally, the decoder $g_\phi(\cdot)$ takes as inputs of the masked state embeddings $\tilde{\eta}_s$ and the actions $\{a_i\}_{i=t}^{t+F-1}$, both of which can be taken from the replay buffer.
The actions are firstly embedded as $d$-dimensional tokens $\{\rva_i\}_{i=t}^{t+F-1}$ with linear layers. 
Then, state and action embeddings are summed with positional encodings \cite{vaswani2017attention} to obtain positional information, and ordered alternatively to form a state-action sequence:
\begin{equation}
\label{eq::sar_input}
    \Tilde{\eta}_{s,a} \coloneqq \textrm{Flat.}(\left\{\Tilde{\rvs}_i, {\rva}_i\right\}_{i=t}^{t+F-1}) + \textrm{Flat.}(\left\{\rvp_i, \rvp_i\right\}_{i=t}^{t+F-1}),
\end{equation}
where $\Tilde{\eta}_{s,a} \in \sR^{2S \times d}$ is the input to the Transformer layers, $\rvp_i \in \sR^{d}$ is the $i^{th}$ positional encoding, and Flat.~denotes the flatten operation.
Then, we gather outputs at the state indices from the Transformer layers, and use a Multi-Layer Perceptron (MLP)-based projection head to obtain the learned representations. 
The causality is enforced through masked self-attention within each Transformer layer.
Let ${\eta}_q$ denote the \textit{query state embeddings}.
We have:
\begin{equation}
\label{eq::tks}
    {\eta}_q \coloneqq \left\{\hat{\rvs}_i \right\}_{i=t}^{t+F-1} = g_\phi(\Tilde{\eta}_{s,a}).
\end{equation}

\subsection{Temporal Contrastive Learning}

\begin{figure}[t]
    \centering
    \includegraphics[width=0.6\linewidth]{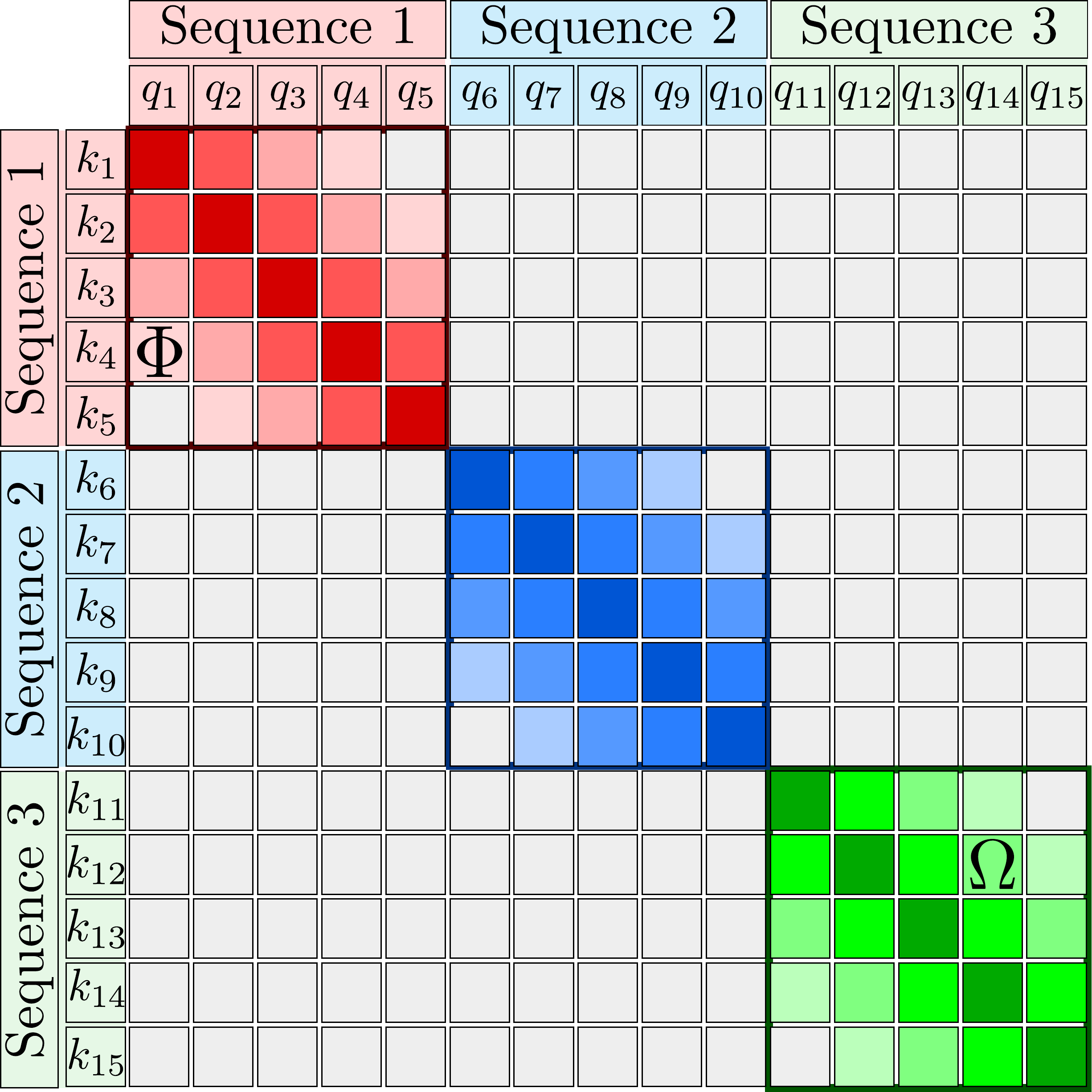}
    \caption{Illustration of the temporal contrastive objective.
    This mock setup contains $3$ sampled sequences with $15$ query-key pairs in total (observation length is $F=5$; batch size is $3$), and models four similarity levels with $L=3$.
    If embeddings are learned from the same sequence, they share the same color scheme.
    The temporal contrastive objective aims to capture a ranked order of state similarities, indicated by the diminishing color intensity from the main diagonal to the off-diagonal cells.
    In this example, $\mathrm{\Phi} = \mathrm{sim(\rvq_1, \rvk_4)} = \mathrm{sim(\rvq, \rvk_{\Delta=3})}$, and $\mathrm{\Omega} = \mathrm{sim(\rvq_{14}, \rvk_{12})} = \mathrm{sim(\rvq, \rvk_{\Delta=2})}$.
    The gray cells denote learned similar scores between $\rvq$ and $\rvk'$, \ie, query-key pairs either belonging to different sampled sequences, or have temporal distance larger than 3.
    These pairs belong to the lowest similarity level.}
    \label{fig:conts_loss}
\end{figure}

The guiding principle of \name~is to learn state representations that evolve temporally in a gradual, smooth fashion, similar to the slowness and variability principles firstly proposed in \cite{jonschkowski2017pves}.
Recall that $\eta_q = g_\phi(\Tilde{\eta}_{s,a})$, $\eta_k = f_{\bar{\theta}}(\eta_o)$ are the query and key trajectories encoded from $\eta_o$, respectively.
In addition, let $\{\eta'_k\} = f_{\bar{\theta}}(\{\eta'_o\})$ be the set of key trajectories encoded from other observation sequences of the same batch, \ie, $\eta_k \notin \{\eta'_k\}$. 
Then, for any query $\rvq \in \eta_q$, we can form its corresponding sets of ranked keys $\{\gK_{\mathrm{\Delta}=l}\}_{l=0}^{L}$, to encourage $\rvq$ is more similar to its temporally adjacent neighbors than those further apart.
That is:
\begin{gather}\raisetag{13pt}\begin{aligned}
& \textrm{sim}(\rvq, \rvk_{\mathrm{\Delta}=0}) > \textrm{sim}(\rvq, \rvk_{\mathrm{\Delta}=1}) > \cdots > \textrm{sim}(\rvq, \rvk_{\mathrm{\Delta}=L}) > \\
& \textrm{sim}(\rvq, \rvk'), \forall \rvk_{\mathrm{\Delta}=l} \in \gK_{\mathrm{\Delta}=l}, \rvk' \in \{\eta'_k\} \cup \gK_{\mathrm{\Delta}>l},
\end{aligned}\end{gather}

where $\rvk_{\mathrm{\Delta}=l} \in \eta_k$ denotes key states that are $l$ units temporally away from $\rvq$, $\rvk' \in \{\eta'_k\}$ are key states that do not come from $\eta_k$, and $L$ is the temporal window size on which the contrastive objective focuses.
\Cref{fig:conts_loss} illustrates this pattern.

\begin{table*}[t]\footnotesize
\centering
\begin{tabular}{ccccccccc}
\toprule 
100k Step Scores & Dreamer & SAC+AE & CURL & DrQ & PlayVirtual & MLR & \textit{Base} & \cellcolor{gray} \name \\
\hline 
Finger, spin & 341 ± 70 & 740 ± 64 & 767 ± 56 & 901 ± 104 & \textbf{915 ± 49} & 907 ± 58 & 853 ± 112 & \cellcolor{gray} 822 ± 6\\
Cartpole, swingup & 326 ± 27 & 311 ± 11 & 582 ± 146 & 759 ± 92 & 816 ± 36 & 806 ± 48 & 784 ± 63 & \cellcolor{gray} \textbf{873 ± 1}\\
Reacher, easy & 314 ± 155 & 274 ± 14 & 538 ± 233 & 601 ± 213 & 785 ± 142 & 866 ± 103 & 593 ± 118 & \cellcolor{gray} \textbf{969 ± 7}\\
Cheetah, run & 235 ± 137 & 267 ± 24 & 299 ± 48 & 344 ± 67 & 474 ± 50 & 482 ± 38 & 399 ± 80 & \cellcolor{gray} \textbf{506 ± 15}\\
Walker, walk & 277 ± 12 & 394 ± 22 & 403 ± 24 & 612 ± 164 & 460 ± 173 & 643 ± 114 & 424 ± 281 & \cellcolor{gray} \textbf{798 ± 42}\\
Ball in cup, catch & 246 ± 174 & 391 ± 82 & 769 ± 43 & 913 ± 53 & 926 ± 31 & 933 ± 16 & 648 ± 287 & \cellcolor{gray} \textbf{944 ± 30}\\
\hline
Mean & 289.8 & 396.2 & 559.7 & 688.3 & 729.3 & 772.8 & 616.8 & \cellcolor{gray} \textbf{818.6}\\
Median & 295.5 & 351.0 & 560.0 & 685.5 & 800.5 & 836.0 & 620.5 & \cellcolor{gray} \textbf{847.5} \\
\midrule
500k Step Scores & & & & & & & & \\
\hline 
Finger, spin & 796 ± 183 & 884 ± 128 & 926 ± 45 & 938 ± 103 & 963 ± 40 & 973 ± 31 & 944 ± 97 & \cellcolor{gray} \textbf{977 ± 8}\\
Cartpole, swingup & 762 ± 27 & 735 ± 63 & 841 ± 45 & 868 ± 10 & 865 ± 11 & 872 ± 5 & 871 ± 4 & \cellcolor{gray} \textbf{878 ± 0} \\
Reacher, easy & 793 ± 164 & 627 ± 58 & 929 ± 44 & 942 ± 71 & 942 ± 66 & 957 ± 41 & 943 ± 52 & \cellcolor{gray} \textbf{977 ± 12} \\
Cheetah, run & 570 ± 253 & 550 ± 34 & 518 ± 28 & 660 ± 96 & \textbf{719 ± 51} & 674 ± 37 & 602 ± 67 & \cellcolor{gray} 712 ± 7\\
Walker, walk & 897 ± 49 & 847 ± 48 & 902 ± 43 & 921 ± 45 & 928 ± 30 & 939 ± 10 & 818 ± 263 & \cellcolor{gray} \textbf{957 ± 22} \\
Ball in cup, catch & 879 ± 87 & 794 ± 58 & 959 ± 27 & 963 ± 9 & 967 ± 5 & 964 ± 14 & 960 ± 10 & \cellcolor{gray} \textbf{974 ± 15} \\
\hline
Mean & 782.8 & 739.5 & 845.8 & 882.0 & 897.3 & 896.5 & 856.3 & \cellcolor{gray} \textbf{912.5} \\
Median & 794.5 & 764.5 & 914.0 & 929.5 & 935.0 & 948.0 & 907.0 & \cellcolor{gray} \textbf{965.5} \\
\bottomrule
\end{tabular}
\caption{Quantitative results for DMC-100k and DMC-500k, as reported in their respective works. \textbf{Bold} values indicate best performance.}
\label{tab:dmc}
\end{table*}

To model such decaying query-key similarities at multiple levels, 
inspired by \cite{hoffmann2022ranking}, we use the InfoNCE loss shown in \cref{eq::infonce} in a recursive manner from $l=0$ to $l=L$.
Specifically, at the $l^{th}$ temporal distance level, the corresponding loss treats $\rvk_{\mathrm{\Delta}=l}$ as positive keys, while the negatives consist of (1) keys from the same trajectory that are temporally further away and (2) keys from other trajectories in the batch.
Formally, let $\gL_{\texttt{MOOSS}} = \sum_{l=0}^{L}\gL_{\rvq}^{l}$ denote \name's objective for query $\rvq$, where $\gL_{\rvq}^{l}$ be the $l^{th}$-level temporal contrastive loss.
We have:
\begin{equation}
\label{eq::loss_l}
    \gL_{\rvq}^{l}=-\log \frac{\sum_{\rvk_{\mathrm{\Delta}=l}} \exp(\textrm{sim}(\rvq, \rvk)/\tau_l)}{\sum_{\rvk_{\mathrm{\Delta}\geq l} \cup \rvk' } \exp(\textrm{sim}(\rvq, \rvk)/\tau_l)},
\end{equation}
where $\rvk_{\mathrm{\Delta}\ge l} \in \eta_k$ denotes key states that are more than or equal to $l$-temporally away from $\rvq$, and $\tau_l < \tau_{l+1}$.
\name's similarity score is measured by bilinear product as in previous works \cite{laskin2020curl, stooke2021decoupling} through $\textrm{sim}(\rvq, \rvk) = \rvq^{T}\rmW\rvk$, where $\rmW$ is a learnable weight matrix.

\subsection{Overall Objective}

The temporal contrastive objective $\gL_{\texttt{MOOSS}}$ serves as an auxiliary loss for RL algorithms.
Let $\gL_{\textrm{rl}}$ denote the loss for the base RL algorithm.
The overall learning objective for the visual RL agent with \name~is:
\begin{equation}
\label{eq::loss_overall}
    \gL_{\textrm{total}}=\gL_{\textrm{rl}} + \lambda \gL_{\texttt{MOOSS}},
\end{equation}
where $\lambda$ is a hyper-parameter trading off the main RL loss and \name's temporal contrastive loss.
We note that the proposed predictive decoder $g_\phi(\cdot)$ is only used during training.
During evaluation, only the observation encoder $f_\theta(\cdot)$ is kept to encode raw, unmasked observations to states.

%% file: files/exp.tex
\section{Experiments}

\subsection{Benchmark Environments}
Sample efficiency of \name~is studied on both the continuous control benchmark DeepMind Control Suite (DMC) \cite{tassa2018deepmind} and the discrete control benchmark Atari \cite{bellemare2013arcade}.
For continuous control, 6 tasks from DMC are used following prior works \cite{yu2021playvirtual, yu2022mask}, including \textit{Finger-spin}, \textit{Cartpole-swingup}, \textit{Reacher-easy}, \textit{Cheetah-run}, \textit{Walker-walk} and \textit{Ball in cup-catch}.
Algorithms are evaluated at 100k and 500k environment steps, referred as DMC-100k and DMC-500k.
For discrete control, the Atari-100k benchmark is used \cite{laskin2020curl, yu2022mask}.
It contains 26 Atari games, and performance is evaluated at 100k interaction steps (\ie, 400k environment steps with action repeat of 4) between the game and RL agents.

\begin{table*}[t]\footnotesize
\centering
\resizebox{\linewidth}{!}{
\begin{tabular}{lccccccccccc}
\toprule 
Game & Human & Random & DER & OTR & CURL & DrQ & SPR & PlayVirtual & MLR & \textit{Base} & \cellcolor{gray} \name \\
\hline Alien & 7127.7 & 227.8 & 802.3 & 570.8 & 711.0 & 734.1 & 841.9 & 947.8 & \textbf{990.1} & 678.5 & \cellcolor{gray} 951.1\\
 Amidar & 1719.5 & 5.8 & 125.9 & 77.7 & 113.7 & 94.2 & 179.7 & 165.3 & \textbf{227.7} & 132.8 & \cellcolor{gray} 207.5 \\
 Assault & 742.0 & 222.4 & 561.5 & 330.9 & 500.9 & 479.5 & 565.6 & \textbf{702.3} & 643.7 & 493.3 & \cellcolor{gray} 667.0 \\
Asterix & 8503.3 & 210.0 & 535.4 & 334.7 & 567.2 & 535.6 & 962.5 & 933.3 & 883.7 & 1021.3 & \cellcolor{gray} \textbf{1140.0} \\
 Bank Heist & 753.1 & 14.2 & 185.5 & 55.0 & 65.3 & 153.4 & \textbf{345.4} & 245.9 & 180.3 & 288.2 & \cellcolor{gray} 288.0 \\
 Battle Zone & 37187.5 & 2360.0 & 8977.0 & 5139.4 & 8997.8 & 10563.6 & 14834.1 & 13260.0 & \textbf{16080.0} & 13076.7 & \cellcolor{gray} 11363.3 \\
 Boxing & 12.1 & 0.1 & -0.3 & 1.6 & 0.9 & 6.6 & 35.7 & \textbf{38.3} & 26.4 & 14.3 & \cellcolor{gray} 22.4\\
 Breakout & 30.5 & 1.7 & 9.2 & 8.1 & 2.6 & 15.4 & 19.6 & \textbf{20.6} & 16.8 & 16.7 & \cellcolor{gray} 16.8\\
Chopper Cmd & 7387.8 & 811.0 & 925.9 & 813.3 & 783.5 & 792.4 & 946.3 & 922.4 & 910.7 & 878.7 & \cellcolor{gray} \textbf{1477.0} \\
Crazy Climber & 35829.4 & 10780.5 & 34508.6 & 14999.3 & 9154.4 & 21991.6 & \textbf{36700.5} & 23176.7 & 24633.3 & 28235.7 & \cellcolor{gray} 21093.3 \\
 Demon Attack & 1971.0 & 152.1 & 627.6 & 681.6 & 646.5 & \textbf{1142.4} & 517.6 & 1131.7 & 854.6 & 310.5 & \cellcolor{gray} 904.0 \\
Freeway & 29.6 & 0.0 & 20.9 & 11.5 & 28.3 & 17.8 & 19.3 & 16.1 & 30.2 & \textbf{30.9} & \cellcolor{gray} 20.3 \\
 Frostbite & 4334.7 & 65.2 & 871.0 & 224.9 & 1226.5 & 508.1 & 1170.7 & 1984.7 & 2381.1 & 994.3 & \cellcolor{gray} \textbf{2898.5} \\
 Gopher & 2412.5 & 257.6 & 467.0 & 539.4 & 400.9 & 618.0 & 660.6 & 684.3 & \textbf{822.3} & 650.9 & \cellcolor{gray} 731.4 \\
 Hero & 30826.4 & 1027.0 & 6226.0 & 5956.5 & 4987.7 & 3722.6 & 5858.6 & 8597.5 & 7919.3 & 4661.2 & \cellcolor{gray} \textbf{9531.2} \\
 Jamesbond & 302.8 & 29.0 & 275.7 & 88.0 & 331.0 & 251.8 & 366.5 & 394.7 & \textbf{423.2} & 270.0 & \cellcolor{gray} 326.3\\
 Kangaroo & 3035.0 & 52.0 & 581.7 & 348.5 & 740.2 & 974.5 & 3617.4 & 2384.7 & \textbf{8516.0} & 5036.0 & \cellcolor{gray} 6122.7 \\
 Krull & 2665.5 & 1598.0 & 3256.9 & 3655.9 & 3049.2 & 4131.4 & 3681.6 & 3880.7 & 3923.1 & 3571.3 & \cellcolor{gray} \textbf{4195.9}\\
 Kung Fu Master & 22736.3 & 258.5 & 6580.1 & 6659.6 & 8155.6 & 7154.5 & 14783.2 & 14259.0 & 10652.0 & 10517.3 & \cellcolor{gray} \textbf{19402.3} \\
 Ms Pacman & 6951.6 & 307.3 & 1187.4 & 908.0 & 1064.0 & 1002.9 & 1318.4 & 1335.4 & \textbf{1481.3} & 1320.9 & \cellcolor{gray} 1362.2 \\
 Pong & 14.6 & -20.7 & -9.7 & -2.5 & -18.5 & -14.3 & -5.4 & -3.0 & \textbf{4.9} & -3.1 & \cellcolor{gray} -4.14 \\
 Private Eye & 69571.3 & 24.9 & 72.8 & 59.6 & 81.9 & 24.8 & 86.0 & 93.9 & \textbf{100.0} & 93.3 & \cellcolor{gray} \textbf{100.0} \\
 Qbert & 13455.0 & 163.9 & 1773.5 & 552.5 & 727.0 & 934.2 & 866.3 & \textbf{3620.1} & 3410.4 & 553.8 & \cellcolor{gray} 3398.0 \\
 Road Runner & 7845.0 & 11.5 & 11843.4 & 2606.4 & 5006.1 & 8724.7 & 12213.1 & 13429.4 & 12049.7 & 12337.0 & \cellcolor{gray} \textbf{19077.0} \\
 Seaquest & 42054.7 & 68.4 & 304.6 & 272.9 & 315.2 & 310.5 & 558.1 & 532.9 & \textbf{628.3} & 471.9 & \cellcolor{gray} 455.5 \\
Up N Down & 11693.2 & 533.4 & 3075.0 & 2331.7 & 2646.4 & 3619.1 & \textbf{10859.2} & 10225.2 & 6675.7 & 4112.8 & \cellcolor{gray} 6963.9\\
\hline
 \textbf{Interquartile Mean} & 1.000 & 0.000 & 0.183 & 0.117 & 0.113 & 0.224 & 0.337 & 0.374 & 0.432 & 0.292 & \cellcolor{gray} \textbf{0.433}\\
 \textbf{Optimality Gap} & 0.000 & 1.000 & 0.698 & 0.819 & 0.768 & 0.692 & 0.577 & 0.558 & \textbf{0.522} & 0.614 & \cellcolor{gray} 0.524 \\
\bottomrule
\end{tabular}}
\caption{Quantitative results for Atari-100k. The best results are highlighted in bold.}
\label{tab:atari}
\end{table*}

\subsection{Baselines and Metrics}
For DMC, \name~is compared with sample-efficient RL methods tailored to continuous control, including Dreamer \cite{hafner2019dream}, SAC+AE \cite{yarats2021improving}, CURL \cite{laskin2020curl}, DrQ \cite{kostrikov2020image}, PlayVirtual \cite{yu2021playvirtual} and MLR \cite{yu2022mask}.
Following previous works, per-task mean (with standard deviation) over $10$ episodic runs with different seeds are reported.
We also report the overall mean and median scores to reflect the general performance.
For Atari experiments, \name~is compared with DER \cite{van2019use}, OTR \cite{kielak2019recent}, CURL \cite{laskin2020curl}, DrQ \cite{kostrikov2020image}, SPR \cite{schwarzer2020data}, PlayVirtual \cite{yu2021playvirtual} and MLR \cite{yu2022mask}. 
Each Atari game is evaluated through $100$ episodic runs across $3$ random seeds following \cite{yu2022mask}.
We leverage the Interquartile Mean (IQM) and the Optimality Gap (OG) metrics with percentile Confidence Intervals (CIs) proposed in Rliable \cite{agarwal2021deep} to study \name's sample efficiency on Atari.
As Atari games are highly non-deterministic with high variances across different games and runs, these aggregate metrics can provide a more rigorous and robust evaluation on algorithmic performance that raw scores.
We report these aggregate metrics alongside individual game scores on Atari-100k with $95\%$ CIs.

\subsection{Implementation}
SAC \cite{haarnoja2018soft} and Rainbow \cite{hessel2018rainbow} are used as continuous and discrete RL algorithms on DMC and Atari environments, respectively.
Following previous works \cite{yu2022mask}, data augmentation including random crop and random intensity are employed as they are proved helpful \cite{laskin2020reinforcement, kostrikov2020image} in improving sample efficiency of RL algorithms.
Based on these, \textit{Base} models \cite{yu2022mask} are firstly devised, which only rely on $\gL_{\textrm{rl}}$ for policy updates by setting $\lambda=0$.
Then, we integrate \name~into the \textit{Base} models.
For all DMC and Atari experiments, we set $\lambda=0.1$ to balance $\gL_{\textrm{rl}}$ and $\gL_{\textrm{\name}}$.
By default, we set the temporal window size $L = 6$ and the mask ratio $p_m = 50\%$, and these key hyper-parameters are further studied in the supplementary material.
More implementation details are also provided in the supplementary material.

\subsection{Comparison with \textit{Base} and State of the Art}

\paragraph{DMC.} We first compare \name~with state-of-the-art visual RL methods and its \textit{Base} model on DMC-100k and DMC-500k. 
The evaluation results are summarized in \cref{tab:dmc}.
From the table, we first observe that \name~consistently improves the performance of its corresponding \textit{Base} model on all tasks by large margins on both benchmarks.
In particular, \name~achieves relative improvements of \textbf{33\%} in mean scores and \textbf{37\%} in median scores on DMC-100k, and \textbf{7\%} in mean scores and \textbf{6\%} in median scores on DMC-500k, respectively.
These improvements clearly demonstrate \name's ability in improving sample efficiency of visual RL algorithms on continuous control tasks.
Second, \name-equipped RL agents outperform previous state-of-the-art methods.
For both DMC-100k and DMC-500k, \name~secures the top performance in five out of six tasks, and obtain the best mean and median scores.
These results indicate that \name~is effective in both sample efficiency and asymptotic performance.

\paragraph{Atari.}
In \cref{tab:atari}, we summarize \name's quantitative results on Atari-100k.
From the table, we again observe that \name~significantly improves the performance of its corresponding \textit{Base} model, having a $\textbf{48\%}$ relative improvement on IQM and a $\textbf{15\%}$ relative improvement on OG, respectively. 
This indicates \name~can greatly improve sample efficiency of visual RL algorithms on discrete control tasks.
In addition, \name~also performs competitively with the current state-of-the-art method MLR, achieving the best IQM score and the second best OG score.
These results demonstrate that \name~has the highest sample efficiency and performs close to human-level performance.

\subsection{Ablation Study}

\begin{table*}[t]\footnotesize
\centering
\begin{tabular}{c | c c c c c c | c c}
\hline Model Variants \textbackslash Task & Finger & Cartpole & Reacher & Cheetah & Walker & Ball & Mean & Median \\
\hline 
\textit{Base}, $\lambda=0$ & \textbf{853 ± 112} & 784 ± 63 & 593 ± 118 & 399 ± 80 & 424 ± 281 & 648 ± 287 & 616.8 & 620.5 \\
$L=0, p_m = 0$ & 829 ± 9 & 795 ± 1 & 702 ± 409 & 401 ± 49 & 68 ± 41 & 766 ± 190 & 593.3 &	734.0 \\
$L=6, p_m = 0$ & 840 ± 20 & 870 ± 1 & 873 ± 291 & 491 ± 11 & 52 ± 24 & 931 ± 35 & 800.9 & \textbf{871.5} \\
$L=6, p_m = 50\%$ as \cite{yu2022mask} & 656 ± 5 & 862 ± 9 & 676 ± 435 & 454 ± 53 & 547 ± 91 & 930 ± 35 & 687.4 & 666.0 \\
\rowcolor{gray}
\name & 822 ± 6 & \textbf{873 ± 1} & \textbf{969 ± 7} & \textbf{506 ± 15} & \textbf{798 ± 42} & \textbf{944 ± 30} & \textbf{818.6} & 847.5 \\
\hline
\end{tabular}
\caption{Ablation on \name's general framework components.} 
\label{tab:abla_gene}
\end{table*}

In this section, we conduct an ablation analysis on DMC-100k to investigate how different design choices of \name~affect its efficacy in improving sample efficiency.
All ablation results are obtained through 10 evaluation runs across different seeds.
Additional ablations are provided in the supplementary material. 

\paragraph{General Framework Components.}

\name~enhances RL algorithms through its (1) temporal contrastive objective facilitated by the (2) random walk-based spatial-temporal masking.
We first evaluate the individual contributions of these components to \name’s performance.
Specifically, in addition to \name, we test four variants of our framework:
(1) First, as previously mentioned, the \textit{Base} model does not rely on $\gL_{\texttt{MOOSS}}$ updates.
(2) We then introduce the contrastive objective into the \textit{Base} model without masking ($p_m = 0$). 
At the same time, we set $L=0$ such that the model does not consider temporally adjacent states thus does not model state evolution.
(3) Next, we improve upon the second model by leveraging the temporal contrastive objective ($L = 6$), while keeping the masking ratio to $0$.
(4) In the fourth variant, we additionally leverage masking with $p_m = 50\%$.
However, instead of doing random walk-based spatial-temporal masking, we apply cube masking \cite{yu2022mask}, which masks the observation cubes uniformly.

Through analysing the results presented in \cref{tab:abla_gene}, we have the following observations:
(1) Both the temporal contrastive objective and the spatial-temporal masking technique improve the sample efficiency of RL algorithms.
All variants equipping $\gL_{\texttt{MOOSS}}$ perform better than the \textit{Base} model in terms of mean and median scores.
(2) The temporal contrastive objective is essential to \name, as it brings a mean score improvement of 35\% and a median score improvement of 19\% when masking is not applied.
(3) Masking is important to the performance of \name~on certain tasks.
We observe that if masking is not used, the \textit{Walker} task shows inferior performance even compared with the \textit{Base} model.
(4) \name~achieves superior performance compared to the \textit{Base} model and its variants on most tasks, having the best mean score performance and the second best median score performance.
This indicates the integration of temporal contrastive objective and the spatial-temporal masking technique can enhance RL agent's understanding of the environment.

\begin{table}[t]	
\centering
\resizebox{\linewidth}{!}{
\begin{tabular}{cccccc}
\hline 
Task & \textit{Base} & \name-NoTrans & \name-S & \name-SAR & \cellcolor{gray} \name \\
\hline 
Finger & 853 ± 112 & \textbf{975 ± 6} & 938 ± 10 & 827 ± 16 & \cellcolor{gray} 822 ± 6 \\
Cartpole & 784 ± 63 & 837 ± 2 & 527 ± 19 & 790 ± 9 & \cellcolor{gray} \textbf{873 ± 1}\\
Reacher & 593 ± 118 & 778 ± 387 & 872 ± 286 & 683 ± 441 & \cellcolor{gray} \textbf{969 ± 7} \\
Cheetah & 399 ± 80 & 427 ± 5 & 543 ± 19 & \textbf{559 ± 7} & \cellcolor{gray} 506 ± 15 \\
Walker & 424 ± 281 & 670 ± 120 & 284 ± 107 & 701 ± 63 & \cellcolor{gray} \textbf{798 ± 42} \\
Ball & 648 ± 287 & \textbf{956 ± 17} & 888 ± 58 & 899 ± 74 & \cellcolor{gray} 944 ± 30\\
\hline
Mean & 616.8 & 773.7 & 675.4 & 743.2 & \cellcolor{gray} \textbf{818.6} \\
Median & 620.5 & 807.5 & 707.5 & 745.5 & \cellcolor{gray} \textbf{847.5} \\
\hline
\end{tabular}}
\caption{Ablation on \name's predictive decoder $g_\phi(\cdot)$.}
\label{tab:abla_dec}
\end{table}

\paragraph{Decoder Setups.}
During training, \name~utilizes an additional predictive decoder $g_\phi(\cdot)$ to generate query states.
We investigate different design choices of $g_\phi(\cdot)$:
(1) \name-NoTrans indicates no Transformer layers are used in the decoder. 
The masked state embeddings $\tilde{\eta}_s$ are only decoded via an MLP head.
(2) For the \name-S case, only state embeddings are used as inputs to the Transformer-based decoder.
(3) \name-SAR indicates states, actions and rewards are all used as inputs to the decoder for query generation.
From the results summarized in \cref{tab:abla_dec}, we confirm that using states and actions as the inputs to \name's predictive decoder provides the best overall mean and median performance scores.
This indicates the meaningful guidance provided by action signals in modeling state evolution across time.
We also observe that \name~stays competitive on most tasks even without the predictive decoder.
This suggests that the core principle of \name~-- to capture the essential dynamics of states by modeling their evolution across time -- is robust and effective. 

%% file: files/conc.tex
\section{Conclusion}
\label{sec::conclusion}

In this work we present~\name, a novel framework with a self-supervised auxiliary objective to improve sample efficiency of visual RL algorithms.
Facilitated by a graph-based spatial-temporal masking approach, \name's temporal contrastive objective goes beyond the binary distinction between positive and negative samples, modeling multiple levels of state similarities across the temporal dimension.
In this way, we encourage the observation encoder to focus on the smoothly evolving nature of state changes over various temporal distances.
The results obtained from extensive experiments and analyses confirm that \name~achieves significant sample efficiency gains over the base method and state-of-the-art works on both DMControl and Atari benchmarks, demonstrating the efficacy of our method.

%% file: files/ack.tex
\small{
\paragraph{Acknowledgements:}
This work is supported in part by Navy N00014-19-1-2373, the joint NSF-USDA CPS Frontier project CNS \#1954556, USDA-NIFA \#2021-67021-34418, and Agriculture and Food Research Initiative (AFRI) grant no. 2020-67021-32799/project accession no.1024178 from the USDA National Institute of Food and Agriculture: NSF/USDA National AI Institute: AIFARMS. 
Work is supported in part by NSF MRI grant \#1725729 \cite{10.1145/3311790.3396649}.
Work also used Delta GPU at NCSA Delta through allocation CIS230331 from the Advanced Cyberinfrastructure Coordination Ecosystem: Services \& Support (ACCESS) program \cite{boerner2023access}, which is supported by NSF grants \#2138259, \#2138286, \#2138307, \#2137603, and \#2138296.
}

%% file: files/app.tex
\appendix
\renewcommand{\thefigure}{A.\arabic{figure}}
\renewcommand{\thetable}{A.\arabic{table}}
\setcounter{figure}{0}
\setcounter{table}{0}
\renewcommand{\theequation}{A.\arabic{equation}}
\setcounter{equation}{0}

\section{Additional Backgrounds}
\label{subsec::back}

\subsection{Soft Actor Critic}
Soft Actor-Critic (SAC) \cite{haarnoja2018soft} is an off-policy, model-free actor-critic Reinforcement Learning (RL) algorithm that follows the entropy-regularized RL framework. This framework introduces the concept of entropy into the RL objective to encourage exploration.
In particular, SAC tries to learn (1) a soft Q-function $Q_\omega(\cdot)$, (2) a soft state value function $V_\psi(\cdot)$, and (3) a policy $\pi_\eta(\cdot)$.
Let $s_t \in \gS$ denote the state at timestep $t$. 
$V_\psi(\cdot)$ is trained to minimize the MSE:
\begin{gather}\raisetag{13pt}\begin{aligned}
    J_V(\psi) = & \mathbb{E}_{s_t \sim \mathcal{D}}[\frac{1}{2}(V_\psi(s_t)-\\
    & \mathbb{E}[Q_w(s_t, a_t)-\log \pi_\eta(a_t | s_t)])^2],
\end{aligned}\end{gather}

where $\mathcal{D}$ is the replay buffer.
$Q_\omega(\cdot)$ is trained to minimize the soft Bellman residual:
\begin{gather}\raisetag{13pt}\begin{aligned}
    J_Q(\omega)= & \mathbb{E}_{(s_t, a_t) \sim \mathcal{D}}[\frac{1}{2}(Q_\omega(s_t, a_t)-\\
    & (r_t+\gamma \mathbb{E}_{s_{t+1} \sim \rho_\pi(s)}[V_{\bar{\psi}}(s_{t+1})]))^2],
\end{aligned}\end{gather}
where $\rho_\pi(s)$ denotes state marginal of the state distribution induced by $\pi$, and $V_{\bar{\psi}}$'s parameters $\bar{\psi}$ are updated by the Exponential Moving Average (EMA) of $\psi$ (or only gets updated periodically) for training stability.
Policy $\pi$ is optimized to maximize the expected return and the entropy at the same time:
\begin{gather}\raisetag{13pt}\begin{aligned}
J_\pi(\eta)= & \mathbb{E}_{s_t \sim \mathcal{D}, \epsilon_t \sim \mathcal{N}}[\log \pi_\eta(f_{\pi_{\eta}}(\epsilon_t ; s_t) | s_t)-\\
&Q(s_t, f_{\pi_{\eta}}(\epsilon_t ; s_t))],
\end{aligned}\end{gather}
where $\epsilon_t$ is the input noise vector sampled from a standard Gaussian $\mathcal{N}$, and $f_{\pi_{\eta}}(\epsilon_t ; s_t)$ denotes actions sampled stochastically from $\pi_\eta(\cdot)$. 
This sampling procedure is accomplished via the reparameterization trick proposed in \cite{DBLP:journals/corr/KingmaW13}.
Given its performance, SAC serves as a robust baseline for continuous control tasks. 

\subsection{Deep Q-Network and Rainbow}

Deep Q-Network (DQN) \cite{mnih2013playing} is the first deep RL algorithm that successfully learns control policies directly from visual data, \ie, image-based observations.
Facilitated by deep neural networks, it greatly improves the training procedure of Q-learning by using (1) an experience replay buffer for drawing samples and (2) a target Q-network $Q_{\omega'}(\cdot)$ to stabilize training.
$Q_{\omega'}(\cdot)$ shares the same architecture with the Q-network $Q_{\omega}(\cdot)$ and is kept frozen as the optimization target every $C$ steps, where $C$ is a hyper-parameter.
$Q_{\omega}(\cdot)$ is trained to minimize the mean square error:
\begin{gather}\raisetag{13pt}\begin{aligned}
J_Q(\omega)=&\mathbb{E}_{(s_t, a_t, s_{t}) \sim \mathcal{D}}[Q_\omega(s_t, a_t)-\\
&(r_t+\gamma \max _{a} Q_{\omega'}(s_{t+1}, a))^2].
\end{aligned}\end{gather}
Rainbow \cite{hessel2018rainbow} is an enhanced DQN variant that amalgamates multiple advancements into a unified RL agent, featuring (1) double DQN \cite{van2016deep}, (2) prioritized experience replay \cite{schaul2015prioritized}, (3) dueling networks \cite{wang2016dueling}, (4) multi-step return \cite{sutton2018reinforcement}, (5) distributional RL as in \cite{bellemare2017distributional}, and (6) noisy layers \cite{fortunato2017noisy}.
By integrating these techniques, Rainbow is considered a robust baseline for discrete control tasks.

\section{\name~Implementation Details}
\label{subsec::impl}

\subsection{Network Architecture}
\name-equipped RL framework consists of two parts: (1) Modules that are necessary for the RL algorithms (SAC and Rainbow), such as the Q-network $Q_{\omega}(\cdot)$ and the observation encoder $f_\theta(\cdot)$; (2) Additional modules required by \name, \ie, the predictive decoder $g_\phi(\cdot)$.

For the first part, we mainly adopt the implementations of SAC and Rainbow from \cite{yu2022mask} for fair comparisons.
Specifically, the observation encoder $f_\theta(\cdot)$ in SAC is built from 4 convolutional layers with ReLU activations, followed by a projection through a linear layer and normalization.
Note that we use a state representation dimension $d=64$ instead of $50$ to allow multi-head attention on $g_\phi(\cdot)$.
On the other hand, in Rainbow, $f_\theta(\cdot)$ includes 3 convolutional layers with ReLU activations, while the Q-learning heads utilize a multilayer perceptron (MLP) design. 
These observation encoders correspond to the query encoder depicted in Fig. 1 of the main paper, and the key encoder $f_{\bar{\theta}}(\cdot)$ adopts the identical architecture as $f_{\theta}(\cdot)$.

The additional predictive decoder $g_\phi(\cdot)$, necessary for \name, comprises 2 transformer encoder layers, each with 4 attention heads. The causality of $g_\phi(\cdot)$ is enforced using a causal attention mask. Actions $a_t$ are converted into action embeddings $\rva_t \in \sR^d$ via a linear layer, and the positional encodings employed are the standard absolute sinusoidal positional encodings introduced in \cite{vaswani2017attention}.

\subsection{General Learning Settings}

We mainly follow the training pipeline of \cite{yu2022mask} to train \name. 
As such, Adam \cite{kingma2014adam} is used to optimize all trainable parameters, and \name~is trained until reaching the designated maximum agent-environment interaction steps.
The hyper-parameters for DMC and Atari are listed in \cref{tab:hy_dmc} and \cref{tab:hy_atari}, respectively, with the \textbf{bolded} ones being tuned for performance analysis.
Notably, in Atari, few games employ a masking ratio of $p_m = 10\%$ and a temporal window size of $L = 2$ to enhance game performance. 
These games typically feature small, fast-moving objects crucial to success. For instance, \textit{Pong} includes a small ping-pong ball crucial for scoring points, while \textit{Gopher} challenges players to stop fast-moving gophers from eating carrots. 
As discussed in the main paper, for games with fast-moving objects, the high masking ratio of $p_m = 50\%$ can lead to excessive information loss, while an overly long contrastive window, with $L = 6$, may become counterproductive. 
This suggests that a large temporal window might encompass states that are too similar, diminishing the effectiveness of \name~in these scenarios.

\section{Additional Experiments}

\subsection{Performance on Harder Tasks from DMC}

\begin{table}[t]\footnotesize	
\centering
\begin{tabular}{cccc}
\hline 
Steps & Model & Reacher, hard & Walker, run \\
\hline 
100k & \textit{Base} & 341 ± 275 & 105 ± 47 \\
\rowcolor{gray}
100k & \name & \textbf{779 ± 379} & \textbf{164 ± 6} \\
\hline 
500k & \textit{Base} & 669 ± 290 & 466 ± 39 \\
\rowcolor{gray}
500k & \name & \textbf{980 ± 11} & \textbf{509 ± 25} \\
\hline
\end{tabular}
\caption{Results on harder DMC tasks.}
\vspace{-4mm}
\label{tab:abla_dmc_hard}
\end{table}

In \cref{tab:abla_dmc_hard}, we extend our analysis by comparing \name~with its \textit{Base} model on two challenging tasks from DMC: \textit{Reacher-hard} and \textit{Walker-run}. 
These tasks have not been previously utilized to evaluate the sample efficiency of visual RL algorithms. 
The results reveal that \name~consistently enhances the performance on these difficult tasks compared to the \textit{Base} variant, underscoring our method's effectiveness. 
Notably, the performance improvements are more pronounced at 100k steps, which is the low data regime.
This further highlights the benefits of modeling the smooth evolution of states on sample efficiency.

\subsection{Temporal Window Size and Masking Ratio}
\label{sec:mr_tw}

\begin{figure}[t]
\centering
\includegraphics[width=0.9\linewidth]{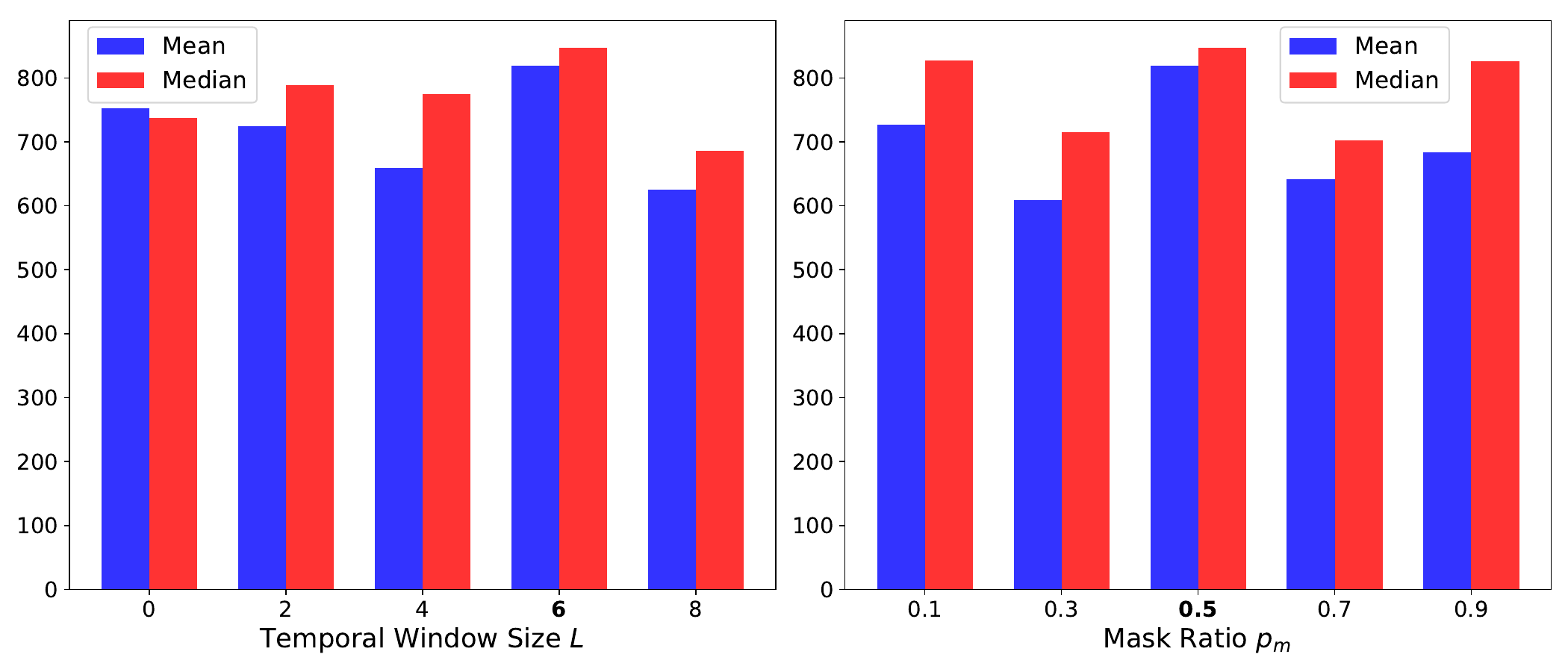}
\caption{Ablation on window size $L$ and masking ratio $p_m$.}
\label{fig:abla_sen}
\end{figure}

In this section, we examine how \name’s key hyper-parameters, \ie, temporal window size $L$ and masking ratio $p_m$, affect its performance.
The results in \cref{fig:abla_sen} on temporal window size present a trend where performance initially fluctuates mildly, reaching a peak, and then deteriorates as the window size expands.
This trend suggests that the context provided by an overly large temporal window can be counterproductive. 
We argue that in the case of a large $L$, for tasks involving repetitive actions (such as \textit{Walker}), states that are temporally distant may also appear similar, leading to confusion and diminishing \name's performance.
We also find that $p_m=50\%$ is a proper choice for \name.
This choice strikes a balance between challenging \name~to exploit spatial-temporal correlations across observations for query generation, and retaining enough unmasked content to facilitate meaningful learning. 
Such level of masking properly ensures that \name~is neither overwhelmed by excessive information loss nor under-stimulated by an abundance of visible data.

\subsection{Ablation on Decoder Depth}

In \cref{tab:abla_depth}, we study the effect of numbers of Transformer layers used in the decoder. 
We observe that the depth of $g_\phi(\cdot)$ is pivotal to \name's performance, with $2$ emerging as the optimal choice. 
The result underscores the necessity of a decoder with balanced power in \name;
it must be sufficiently effective in reducing possible ambiguities in masked state embeddings, but not so dominant as to usurp the learning role of the observation encoder.
\begin{table}[t]\footnotesize	
\centering
\begin{tabular}{cccc}
\hline 
Depth & $g_\phi(\cdot)$ Size & Mean & Median \\
\hline 
1 & 63.27K & 660.1 & 690.0 \\
\rowcolor{gray}
2 (ours) & 113.25K & \textbf{818.6} & \textbf{847.5} \\
3 & 163.24K & 695.8 & 753.5 \\
4 & 213.22K & 667.9 & 847.0 \\
\hline
\end{tabular}
\caption{Ablation on decoder depth.}
\vspace{-4mm}
\label{tab:abla_depth}
\end{table}

\section{Discussion on Limitations} 
While effective, \name's performance gain on Atari is relatively lower compared to DMC.
Delving into this, we observe that \name~does not perform as well in Atari games featuring small, fast-moving objects crucial to success, like bullets. This is particularly evident in games such as Battle Zone, compared to its performance in other games.
This may be because \name's temporal contrastive objective becomes less effective in capturing drastic key changes across states, and is further challenged by spatial-temporal masking, which might result in excessive information loss.
Besides, \name~requires hyper-parameters that may need additional tuning for different applications.

Additionally, we recognize that certain tasks may violate \name's ``gradually evolving state'' assumption, as discussed in the Limitation Section.
However, we first note that in scenarios with frequent background changes (\eg, Hero from Atari), \name~proves \textit{advantageous} as it guides the encoder to filter out task-irrelevant background information, thereby focusing on task-essential elements.
Second, while \name~does not inherently address fast moving agents algorithmically, this issue is mitigated by the {\small\textsf{action\_repeat}} hyperparameter in RL algorithms. 
{\small\textsf{action\_repeat}} is usually adjusted to a small value for environments with rapid observation/agent changes (\eg, 2 for Spin vs.\ 8 for Swing from DMControl), to stabilizes temporal state dynamics and thus enhances RL model performance.
In \name, {\small\textsf{action\_repeat}} is not specifically tuned.
Thus, given \name's benefit from this mechanism, violations of gradual state evolution assumption are likely rare.

\begin{figure*}[t]
\centering
\begin{tabular}{ll}
\toprule
Hyper-parameter & Value \\
\midrule
Frame stack ($c / 3$) & 3 \\
Observation rendering & $(100,100)$ \\
Observation downsampling ($H \times W$) & $(84,84)$ \\
Augmentation & Random crop and random intensity \\
Replay buffer size & 100000 \\
Initial exploration steps & 1000 \\
Action repeat & \begin{tabular}[t]{@{}l@{}}
2 \textit{Finger-spin} and \textit{Walker-walk}; \\
8 \textit{Cartpole-swingup}; \\
4 otherwise
\end{tabular} \\
Evaluation episodes & 10 \\
Optimizer & Adam \\
$\left(\beta_1, \beta_2\right)$ (Except $\alpha$) & $(0.9,0.999)$ \\
$\left(\beta_1, \beta_2\right) \rightarrow(\alpha)$ (temperature in SAC) & $(0.5,0.999)$ \\
Learning rate for base RL modules & \begin{tabular}[t]{@{}l@{}}
0.0002 \textit{Cheetah-run}; \\
0.001 otherwise
\end{tabular} \\
Learning rate for \name-specific modules & \begin{tabular}[t]{@{}l@{}}
0.0001 \textit{Cheetah-run}; \\
0.0005 otherwise
\end{tabular} \\
Learning rate warmup for \name-specific modules & 6000 steps \\
Learning rate & 0.0001 \\
Batch size for policy learning & 512 \\
Batch size for auxiliary task & 128 \\
Q-function EMA $m$ & 0.99 \\
Critic target update frequency & 2 \\
Discount factor & 0.99 \\
Initial temperature & 0.1 \\
Target network update period & 1 \\
Target network EMA $m$ & \begin{tabular}[t]{@{}l@{}}
0.9 \textit{Walker-walk}; \\
0.95 otherwise \\
\end{tabular} \\
State representation dimension $d$ & 64 \\
\midrule
\multicolumn{2}{l}{ \name~Specific Hyper-parameters } \\
\midrule
Weight of \name~loss $\lambda$ & 0.1 \\
Sequence length $F$ & 16 \\
Cube spatial size $h \times w$ & $7 \times 7$ \\
Cube temporal length $f$ & \begin{tabular}[t]{@{}l@{}}
4 \textit{Cartpole-swingup} and \textit{Reacher-easy} \\
8 otherwise
\end{tabular} \\
Initial Contrastive temperature $\tau_0$ & 0.07 \\
Contrastive temperature skip $\tau_{l+1}-\tau_l$ & 0.075 \\
\textbf{Predictive decoder $\mathbf{g_\phi(\cdot)}$ depth} & 2\\
\textbf{Random walk mask ratio $\mathbf{p_m}$} & $50 \%$\\
\textbf{Temporal window size $\mathbf{L}$} & 6 \\
\bottomrule
\end{tabular}
\captionof{table}{Hyper-parameters used for DMC.}
\label{tab:hy_dmc}
\end{figure*}

\begin{figure*}[t]	
\centering
\begin{tabular}{ll}
\toprule
Hyper-parameter & Value \\
\midrule
Gray-scaling & True \\
Frame stack ($c / 3$) & 4 \\
Observation downsampling ($H \times W$) & $(84,84)$ \\
Augmentation & Random crop and random intensity \\
Action repeat & 4 \\
Training steps & $100 \mathrm{k}$ \\
Max frames per episode & $108 \mathrm{k}$ \\ 
Reply buffer size & $100 \mathrm{k}$ \\
Minimum replay size for sampling & 2000 \\
Mini-batch size & 32 \\
Optimizer, (learning rate, $\beta_1, \beta_2, \epsilon$) & Adam, (0.0001, 0.9, 0.999, 0.00015) \\
Max gradient norm & 10 \\
Update & Distributional Q \\
Dueling & True \\
Support of Q-distribution & 51 bins \\
Discount factor & 0.99 \\
Reward clipping Frame stack & {$[-1,1]$} \\
Priority exponent, correction & 0.5, $0.4 \rightarrow 1$ \\
Exploration & Noisy nets \\
Noisy nets parameter & 0.5 \\
Evaluation trajectories & 100 \\
Replay period every & 1 step \\
Updates per step & 2 \\
Multi-step return length & 10 \\
Q-network: channels & $32,64,64$ \\
Q-network: filter size & $8 \times 8,4 \times 4,3 \times 3$ \\
Q-network: stride & $4,2,1$ \\
Q-network: hidden units & 256 \\
Target network update period & 1 \\
EMA coefficient $m$ & 0 \\
\midrule
\multicolumn{2}{l}{ \name~Specific Hyper-parameters } \\
\midrule
Weight of \name~loss $\lambda$ & 0.1 \\ 
Sequence length $F$ & 16 \\
Cube spatial size $h \times w$ & $7 \times 7$ \\
Cube temporal length $f$ & 4 \\ 
Initial Contrastive temperature $\tau_0$ & 0.07 \\
Contrastive temperature skip $\tau_{l+1}-\tau_l$ & 0.075 \\
Predictive decoder $g_\phi(\cdot)$ depth & 2 \\
\textbf{Random walk mask ratio $\mathbf{p_m}$} & \begin{tabular}[t]{@{}l@{}} 
$10 \%$ \textit{Gopher}, \textit{Kangaroo}, \\
\textit{Ms Pacman}, \textit{Pong}, \textit{Seaquest} \\
$50 \%$ otherwise 
\end{tabular} \\
\textbf{Temporal window size $\mathbf{L}$} & \begin{tabular}[t]{@{}l@{}} 
$2$ \textit{Gopher}, \textit{Kangaroo}, \\
\textit{Ms Pacman}, \textit{Pong}, \textit{Seaquest} \\
$6$ otherwise 
\end{tabular} \\
\bottomrule
\end{tabular}
\captionof{table}{Hyper-parameters used for Atari.}
\label{tab:hy_atari}
\end{figure*}